\pdfoutput=1

\documentclass[11pt]{article}

\usepackage{acl}

\usepackage{times}
\usepackage{latexsym}
\usepackage{balance}
\usepackage[T1]{fontenc} 
\usepackage[utf8]{inputenc}
\usepackage{microtype}

\usepackage{xurl}
\usepackage{graphicx} 
\usepackage{makecell}
\usepackage[caption=false]{subfig}
\usepackage{amsmath}
\usepackage{amssymb}
\usepackage{multirow}
\usepackage{enumitem}
\usepackage{color, colortbl}

\definecolor{Green}{RGB}{34,139,34}
\definecolor{Red}{RGB}{238, 044, 044}

\newcommand{\textbfGreen}[1]{\textbf{\color{Green}#1}}
\newcommand{\textRed}[1]{{\color{Red}#1}}

\title{Efficient, Uncertainty-based Moderation of Neural Networks \\ Text Classifiers}

\author{Jakob Smedegaard Andersen \and Walid Maalej\\
         University of Hamburg \\ Hamburg, Germany \\
         \texttt{\{jakob.smedegaard.andersen,walid.maalej\}@uni-hamburg.de}}

\begin{document}
\maketitle
\begin{abstract}
To maximize the accuracy and increase the overall acceptance of text classifiers, we propose a framework for the efficient, in-operation moderation of classifiers' output. Our framework focuses on use cases in which F1-scores of modern Neural Networks classifiers (ca.~90\%) are still inapplicable in practice. 
We suggest a semi-automated approach that uses prediction uncertainties to pass unconfident, probably incorrect classifications to human moderators. 
To minimize the workload, we limit the human moderated data to the point where the accuracy gains saturate and further human effort does not lead to substantial improvements. 
A series of benchmarking experiments based on three different datasets and three state-of-the-art classifiers show that our framework can improve the classification F1-scores by 5.1 to 11.2\% (up to approx.~98 to 99\%), while reducing the moderation load up to 73.3\% compared to a random moderation.
\end{abstract}

\section{Introduction}
Accurately classifying an overwhelming amount of textual data is a common research challenge \cite{pouyanfar2018survey}.
In recent years, machine learning approaches, particularly Neural Networks (NNs), have received great attention to support textual  classification \cite{lai2015recurrent}. 
However, in practice, fully automated approaches are still rare due to the general lack of top, almost-perfect classification accuracy.
If the accuracy of a trained and hyper-tuned state-of-the-art classifier still does not meet the domain requirements, a full manual approach is likely to be the fall-back solution.

To prevent classification mistakes and strengthen the overall acceptance  of artificial decision making, socio-technical approaches that integrate human domain experts in the decision loop are gaining in importance \cite{holzinger2016interactive}.
Recent research has shown that including the prediction uncertainty of NNs can detect more complex text inputs, i.e., either short or very long texts with less informative tokens \cite{xiao2019quantifying}, and probably wrong \cite{hendrycks17baseline} predictions, which are worth checking manually. 
Since human resources are cost intensive and do not scale well to larger workloads, moderation processes should be designed with human-resource-efficiency in mind. 
Yet, the efficient in-operation integration of human efforts for building  semi-automated decision-making systems -- i.e. moderated classifiers -- 
remains largely unexplored.

This paper introduces a novel framework for the efficient moderation of NN text classifiers. Our framework extends a given NN with human expertise to create a semi-automated decision-making system. Text instances are moderated manually when classifier outcomes are likely to be false. 
We use the concept of prediction uncertainty \cite{der2009aleatory} to quantify the reliability of a classification.  
When a classifier is highly uncertain, we let human moderators intervene. 
To minimize human efforts, we propose to limit the moderation to the point where the classification accuracy gain saturates.
While active learning \cite{settles1995active} aims to limit human efforts during the \textit{training} of classifiers, our moderation framework seeks to substantially enhance the accuracy of \textit{trained} and \textit{already deployed} classifiers -- surpassing the maximum achievable accuracy of an automatic classifier to still achieve an almost-perfect in-operation accuracy.

Our contribution is twofold. First, we introduce a novel saturation-based framework for the efficient, in-operation moderation of NN-based text classifiers.
Second, we empirically evaluate the accuracy improvement and needed moderation load for three English text classification tasks including hate speech detection, sentiment analysis, and topic classification. We run multiple training trails using different predictive uncertainty estimation approaches and compare their initial and post-moderation F1-scores and evaluate their suitability.

The remainder of the paper is structured as follows:  
Section \ref{Moderated Classifier} introduces our moderation framework and outlines the uncertainty estimation techniques to decide when to moderate. Then, Section \ref{Evaluation} describes the setting to evaluate our framework.  Afterwards, we report on the experiment results in Section \ref{Results} and discuss the implications and limitations of our findings in Section \ref{dicussion}. Finally, Section \ref{Related Work} discusses related work and Section \ref{end} concludes the paper.

\section{Moderating NN Classifiers}
\label{Moderated Classifier}

\subsection{Moderated Classifiers}
\label{Model}

In order to prevent low confident classifications and increase the accuracy of NN text classifiers, we propose the concept of a moderated classifier. A moderated classifier combines an artificial classifier with a human oracle. The oracle steers the decision-making in case the machine is unable to provide a reliable outcome. 
The level of reliability is measured based on the predictive uncertainty of the artificial classifier.

Misclassifications occur when the inferred label $ y_i $ of an input $ x_i $ does not correspond to the actual true label $\hat{y}_i$ and thus $ \hat{y}_i\neq y_i $ holds. If only classifications made under a high uncertainty are delegated to a human oracle, the number of misclassifications can be significantly reduced while keeping manual workloads low.

A moderated classifier $ f_{mod}^\omega $ is created from an artificial classifier $ f^\omega $ as follows:
\begin{equation}
	f^\omega_{mod}(x) := \begin{cases}
		f^\omega(x) &  \text{if $u[y|x, \omega] \leq \vartheta_u$} \\ 
		o_{H}(x) &\text{else}
	\end{cases}
\end{equation}
where $ o_{H}: X\rightarrow Y $ represents the human oracle, $u[y|x, \omega] \in U \subset \mathbb{R}^+$ an uncertainty measure of $f^\omega(x)$ and $\omega$ the learned parameters of $f$. 
If the uncertainty is below a threshold $\vartheta_u \in U$, the inferred label $y = f^\omega(x)$ is considered to be reliable and will be kept.  
If the threshold is breached ($u[y|x, \omega] > \vartheta_u$), a human oracle $ o_H $ is consulted and his or her decision is deemed correct. 
The oracle can also include a group of moderators to share the workload or increase accuracy in the case of contradictions. Conflicts could, e.g., be solved following inter-annotator agreement approaches \cite{artstein2008inter}. In this paper, we focus on single human moderation.

\subsection{Determining Uncertainty Thresholds}
\label{sec:saturation}

The efficient use of (the usually limited) human resources is essential for semi-automated classification approaches. The moderation of text classification can be particularly time-consuming and cost intensive, as moderators might need to carefully read and think about the text.  
It is thus important to limit the moderation effort to a reasonable and worthwhile amount. 
Limiting the moderation effort is a trade-off between saving resources and increasing accuracy.

We suggest a \textbf{saturation-based} moderation strategy to determine the uncertainty threshold $\vartheta_u$. 
As we assume misclassifications to occur more frequently with high uncertainty scores, the moderation is expected to become less efficient with an increasing moderation load. 
At some point, significant improvements may not be achieved and further efforts have a decreasing impact in terms of increasing accuracy. 
A saturation-based strategy seeks to limit the moderation up to a point, where the expected accuracy improvement turns and becomes less rewarding. 

\begin{figure}[!t]
	\centering
	\includegraphics[width=.40\textwidth]{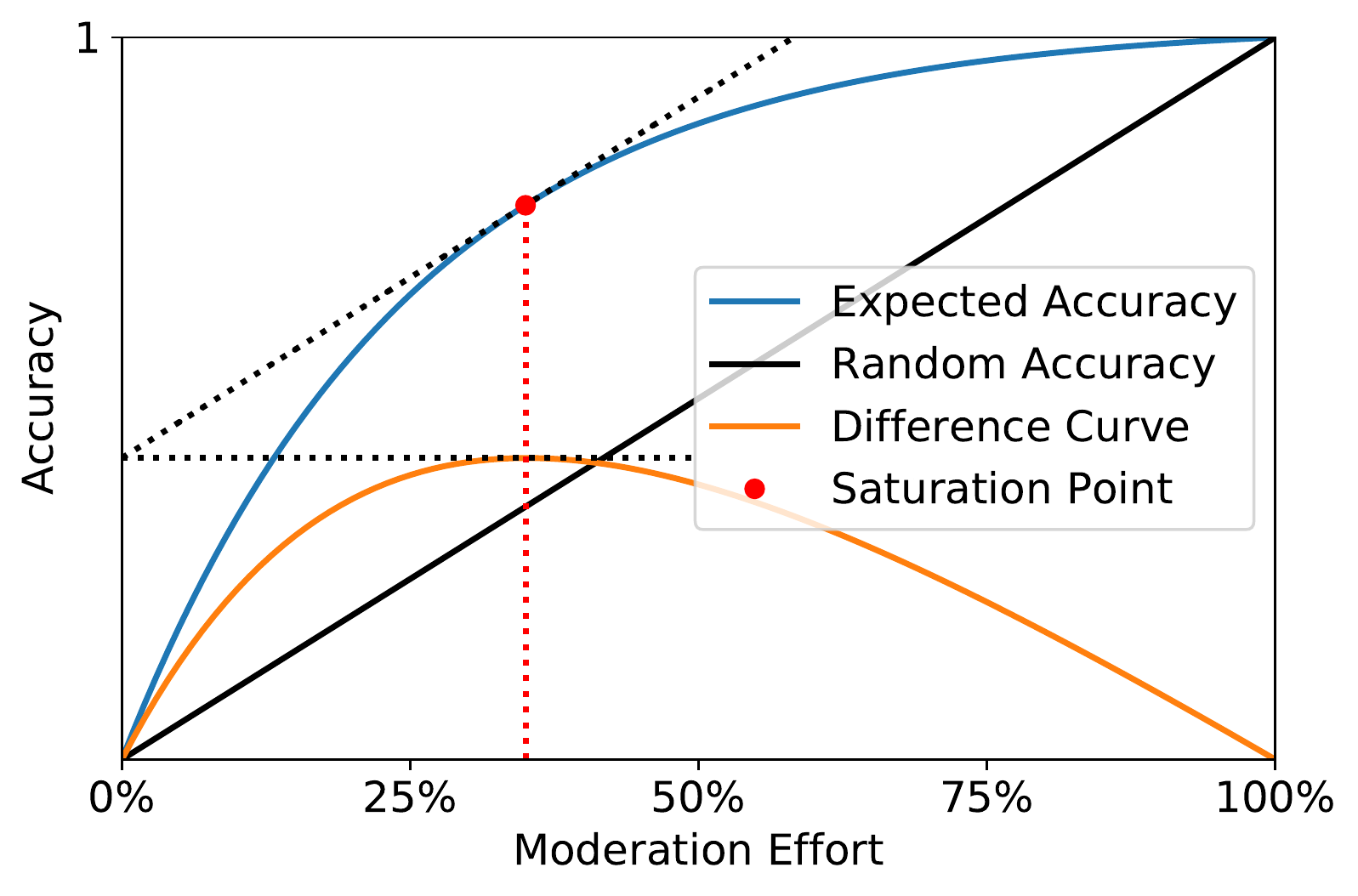}
	\caption{
		Saturation detection for manual moderation.
		Saturation is reached at the highest point of the difference between the expected and random accuracy curve.
	}
	\label{saturation}
\end{figure}

Figure \ref{saturation} shows a hypothetical saturation curve for a moderated classifier. 
The blue curve represents the expected accuracy of our framework when a certain amount of the most uncertain predictions are manually moderated.
The accuracy of a moderated classifier is based on (a) the manual classification and (b) the accuracy of the model classifying the instances which are not passed to a human. An accuracy of 100\% is reached when 100\% of the instances are correctly decided manually. 
The plotted accuracy curve is of shape $f(x)=a (1 - \text{e}^{-bx})$.
The black line shows the  moderation accuracy when instances are randomly sampled for moderation.
A random moderation selects the to-be-moderated instances independently and evenly distributed from a dataset. 
Since a random sample is expected to include the same proportions of misclassifications, the accuracy gain increases linearly over the moderation effort.

A natural \textbf{point of saturation} can be calculated as the highest point of the difference curve between the expected and the random accuracy \cite{satapaa2011knee}. 
It describes the situation where a continued uncertainty-based moderation would become less effective than the moderation of randomly selected instances. We argue that the moderation should be stopped at this natural limit for keeping the manual effort efficient.  

\subsection{Uncertainty Modeling Techniques}
\label{Obtaining Uncertainty1}

Uncertainty in NNs classifications generally occurs when inputs are corrupted by noise or not from the distribution of the training dataset \cite{der2009aleatory}. To estimate model and data uncertainties in $f^\omega$ we use techniques which have performed well in similar uncertainty-based tasks like computer vision \cite{kendall2017what} and active learning \cite{burkhardt2018semi,gal2017deep}. 
We focus on the following uncertainty modeling techniques and on a baseline:

\paragraph{Baseline}
We consider deterministic softmax outcomes of a usual NN as a baseline indicator of confidence (e.g. \citeauthor{hendrycks17baseline} \citeyear{hendrycks17baseline}).
A softmax activation function applied to the last layer of a NN normalizes the network's outcome into pseudo class probabilities. 
\paragraph{Monte Carlo Dropout (MCD)}
According to \citet{gal2016dropout},  Dropout can be interpreted as a Bayesian approximation of a Gaussian process. 
Dropout is generally used as a stochastic regularization technique for NNs to prevent overfitting \cite{srivastava2014dropout}.
To perform approximate Bayesian inference, a NN is  trained with Dropout applied before every weight layer and a softmax activation function after the last layer. 
Then, Dropout is additionally performed at prediction time to sample from an approximated distribution of the real class posterior. 
\paragraph{Bayes by Backprob (BBB)} Bayes by Backprob is another Bayesian approximation technique to model uncertainties in NNs \cite{blundell2015weight}. In BBB, a probability distribution is placed over the NNs weights $\omega$. The approach seeks to learn the posterior distribution $p(\omega|D)$ given the training data $D$. Due to intractabilities, the posterior distribution is approximated by a variational distribution $q(\omega|\theta)$ by minimizing the Kullback–Leibler divergence \cite{kullback1997information}. As for MCD, multiple forward passes are performed to sample over different weights $\hat{\omega}_t \sim q(\omega|\theta)$.

\paragraph{Ensemble}
An Ensemble of multiple independent deterministic NNs is an alternative approach to Bayesian approximation \cite{lakshminarayanan2017simple}. The idea is to use $M$ independent trained NN classifiers and average their softmax outcomes to a single classification score. The parameters of the different models are randomly initialized and individually optimized.

We use score functions based on the uncertainty modeling techniques to quantify the uncertainty of individual classifications \cite{lewis1994sequential,burkhardt2018semi,gal2017deep}. 
Score functions aim to report high uncertainty values for unreliable classifications.
Commonly used metrics are the \textit{Least Confidence} \cite{culotta2005reducing}, \textit{Smallest Margin} \cite{scheffer2001active}, and \textit{Mutual Information} \cite{houlsby2011bayesian}. 

\section{Experimental Design}
\label{Evaluation}
\subsection{Research Questions and Method}

To evaluate our saturation-based moderation framework, we conduct benchmarking experiments using different public datasets and NN classifiers.
We focus on the following research questions: 
\begin{enumerate}[leftmargin=.9cm]
	\item[\textbf{RQ1}]How does uncertainty modeling improve the performance of unmoderated and moderated classifiers?
	\item[\textbf{RQ2}] How much accuracy improvement would the saturation-based moderation bring and at what cost?
\end{enumerate}
With RQ1, we particularly aim to check whether a mere uncertainty modeling (i.e.~still a full automated classification without  moderation) would solve the problem and lead to top classification accuracy close to 99\%. With RQ2, we aim to evaluate our framework's cost/effect in different settings. 

To answer the research questions, we perform a series of machine learning experiments. 
First, we assess the initial performance of three classifiers extended with the different uncertainty modeling techniques. 
We apply the micro F1-score to measure the accuracy of classifiers on the actual classification task. 
Further, we assess a model's ability to detect misclassifications. We compute  the AUC-ROC score, which measures the \textit{Area Under the Receiver Operating Characteristics} curve, based on the correct (positive class) and misclassified (negative class) outcomes. All experiments are performed on the held-out evaluation set. 

Second, to evaluate the moderated classification and determine the best suited uncertainty estimation technique, we examine how a moderation affects the overall F1-score when a certain number of uncertain instances get moderated manually. Further, we calculate the points of saturation to estimate the achievable F1-scores while limiting human moderation efforts.

\begin{table*}[ht]
	
	\centering
	\caption{Effect of extending NN text classifiers with the uncertainty modeling techniques (\textbf{without manual moderation}). Each cell shows the mean $|$ standard deviation of five independent classification runs. For each of the nine experiments (3 classifiers x 3 datasets) the scores of the best performing uncertainty modeling technique are highlighted in green and the lowest scores in red.
	}
	\label{Performance Overview}
	\resizebox{1\textwidth}{!}{%
		\begin{tabular}{c|c|c|c|c||c|c|c|c||c|c|c|c||c|c|c|c}
			\cline{2-13}
			\multicolumn{1}{l|}{} & \multicolumn{4}{c||}{\textbf{Hate Speech}} & \multicolumn{4}{c||}{\textbf{IMDB}} & \multicolumn{4}{c||}{\textbf{20NewsGroups}}  \\ \cline{1-13}
			\multicolumn{1}{|c||}{\textbf{Metrics}}        & Baseline           & MCD         & BBB         & Ensemble            & Baseline           & MCD         & BBB         & Ensemble & Baseline           & MCD         & BBB         & Ensemble    \\ \cline{1-13}
			
			\multicolumn{8}{l}{} \\[-0.7ex] \hline
			
			\multicolumn{1}{|c||}{\textbf{F1-score} $\uparrow$}  & 
			\textRed{90.2}$|$0.1 & \textbfGreen{90.6}$|$0.2 &   90.4$|$0.2  & 90.4$|$0.3  & 
			\textRed{88.7}$|$0.1  & 89.0$|$0.1  & 89.0$|$0.1  & \textbfGreen{89.6}$|$0.2   & 
			\textRed{86.9}$|$0.2  & 87.1$|$0.1  & 87.4$|$0.3  & \textbfGreen{90.1}$|$0.4   & 
			
			\parbox[t]{2mm}{\multirow{5}{*}{\rotatebox[origin=c]{90}{\textbf{CNN}}}} \\ \cline{1-13}
			\Xcline{1-13}{2\arrayrulewidth}
			
			\multicolumn{1}{|c||}{Mean Conf. Mis.}            & 
			89.3$|$0.2 & 83.4$|$0.4  & 87.5$|$0.3  & 85.2$|$0.4  & 
			88.5$|$0.6  & 79.5$|$0.2 & 82.5$|$0.7 & 82.5$|$0.2 & 
			65.4$|$0.7 & 57.5$|$1.9& 57.2$|$0.8  & 55.5$|$0.8 & 
			\\ \cline{1-13}
			
			\multicolumn{1}{|c||}{Mean Conf. Suc.}            & 
			98.3$|$0.1 & 96.9$|$0.1  & 98.1$|$0.1  & 97.6$|$0.1  &
			97.9$|$0.1  & 95.2$|$0.1 & 96.3$|$0.1 & 96.3$|$0.1 & 
			94.4$|$0.3 & 85.1$|$0.4 & 91.6$|$0.2 & 90.8$|$0.2 & 
			\\ \cline{1-13}
			
			\multicolumn{1}{|c||}{Range $\uparrow$} & 
			\textRed{8.7} & \textbfGreen{13.5}  & 10.6 &  12.4 &
			\textRed{9.4}        & \textbfGreen{15.7}       & 13.8      & 13.8      & 
			\textRed{29.0}       & 34.6    & 34.4          & \textbfGreen{35.3} & 
			\\ \Xcline{1-13}{2\arrayrulewidth}
			
			\multicolumn{1}{|c||}{AUC-ROC $\uparrow$}      & 
			86.3$|$0.5 & \textRed{86.1}$|$0.6  & 86.3$|$0.4  & \textbfGreen{86.8}$|$0.3  &
			\textRed{83.7}$|$0.2  & \textbfGreen{84.0}$|$0.1 & 83.8$|$0.4 &83.8$|$0.2 & 
			89.8$|$0.1 & 90.0$|$0.2 & \textbfGreen{90.2}$|$0.4 & \textRed{89.2}$|$0.5&
			\\ \hline
			
			\multicolumn{8}{l}{} \\[-0.7ex] \hline
			
			\multicolumn{1}{|c||}{\textbf{F1-score}}     & 
			\textRed{91.2}$|$0.2 & \textbfGreen{91.4}$|$0.1  & 91.3$|$0.2 & 91.3$|$0.1 &
			\textRed{88.9}$|$0.2 & \textbfGreen{89.7}$|$0.2  & 89.3$|$0.1  & 89.5$|$0.1  & 
			88.2$|$0.3  & 88.7$|$0.2  & \textRed{86.8}$|$0.2  & \textbfGreen{89.5}$|$0.2  
			& \parbox[t]{2mm}{\multirow{5}{*}{\rotatebox[origin=c]{90}{\textbf{KimCNN}}}}\\
			\Xcline{1-13}{2\arrayrulewidth}
			
			\multicolumn{1}{|c||}{Mean Conf. Mis.}            & 
			82.6$|$0.8 & 77.5$|$0.3  & 82.8$|$0.3  & 81.7$|$0.2  &
			78.5$|$0.5 & 73.4$|$0.2  & 82.3$|$0.3  & 76.5$|$0.5 & 
			54.2$|$0.5  & 48.1$|$0.4  & 60.1$|$0.3  & 52.0$|$0.4 & 
			\\ \cline{1-13}
			
			\multicolumn{1}{|c||}{Mean Conf. Suc.}            & 
			97.0$|$0.3 & 95.4$|$0.1  & 97.1$|$0.1  & 96.8$|$0.1  &
			97.5$|$0.1 & 92.2$|$0.1  & 96.5$|$0.1  & 93.6$|$0.2 & 
			90.8$|$0.3  & 84.5$|$0.4  & 92.0$|$0.1  & 89.5$|$0.2 & 
			\\ \cline{1-13}
			
			\multicolumn{1}{|c||}{Range }        &  
			14.4 & \textbfGreen{17.9}  & \textRed{14.3}  & 15.1 &
			15.9      & \textbfGreen{18.8}    & \textRed{14.2}            & 17.1      & 
			36.6       & 36.4      &    \textRed{31.9}   & \textbfGreen{37.5} & 
			\\ 
			\Xcline{1-13}{2\arrayrulewidth}
			
			\multicolumn{1}{|c||}{AUC-ROC }     & 
			\textRed{85.1}$|$0.6 & \textbfGreen{87.6}$|$0.2  & 85.2$|$0.4  & 85.9$|$0.0  &
			\textRed{83.6}$|$0.4 & 84.7$|$0.3  & \textbfGreen{85.0}$|$0.2 & 84.0$|$0.4  & 
			88.4$|$0.3  & \textbfGreen{89.1}$|$0.3  & 88.5$|$0.5  & \textRed{88.2}$|$0.2   & 
			\\ \hline
			
			\multicolumn{8}{l}{} \\[-0.7ex] \hline
			
			\multicolumn{1}{|c||}{\textbf{F1-score} $\uparrow$}     & 
			\textRed{94.0}$|$0.2 & \textbfGreen{94.1}$|$0.1  & -  & \textRed{94.0}$|$0.1  &
			\textRed{93.7}$|$0.1 & \textRed{93.7}$|$0.1  & -  & \textbfGreen{93.9}$|$0.2  & 
			90.5$|$0.4 & \textRed{90.4}$|$0.4  & -  & \textbfGreen{91.1}$|$0.3  & 
			\parbox[t]{2mm}{\multirow{5}{*}{\rotatebox[origin=c]{90}{\textbf{DistilBERT}}}}  \\ 
			\Xcline{1-13}{2\arrayrulewidth}
			
			\multicolumn{1}{|c||}{Mean Conf. Mis.}            & 
			86.6$|$0.5 & 83.3$|$0.6  & -  & 85.8$|$1.2  &
			85.7$|$1.1 & 82.1$|$0.8  & -  & 82.8$|$0.8  & 
			71.1$|$1.7 & 66.4$|$0.9  & -  & 68.3$|$0.8  & 
			\\ \cline{1-13}
			
			\multicolumn{1}{|c||}{Mean Conf. Suc.}            & 
			98.5$|$0.1 & 98.1$|$0.1  & -  & 98.5$|$0.1  &
			98.2$|$0.3 & 97.5$|$0.2  & -  & 97.7$|$0.2  & 
			95.1$|$0.2 & 93.5$|$0.2  & -  & 94.5$|$0.1  & 
			\\ \cline{1-13}
			
			\multicolumn{1}{|c||}{Range $\uparrow$}        & 
			\textRed{11.9} & \textbfGreen{14.8} &  - & 12.7 &
			\textRed{12.5} & \textbfGreen{15.4} & - & 14.9 &
			\textRed{24.0} & \textbfGreen{27.1} & - &  26.2& 
			\\ \Xcline{1-13}{2\arrayrulewidth}
			
			\multicolumn{1}{|c||}{AUC-ROC $\uparrow$}     & 
			\textRed{89.5}$|$0.5 & \textbfGreen{91.6}$|$0.3  & -  & 91.4$|$0.4  &
			\textRed{88.7}$|$0.4 & 88.9$|$0.4  & -  & \textbfGreen{89.0}$|$0.3  & 
			\textRed{90.2}$|$0.4 & \textbfGreen{90.4}$|$0.3  & -  & \textbfGreen{90.4}$|$0.3   &
			\\ \hline
			
		\end{tabular}
	}
	
	\label{t:performance}
\end{table*}

\subsection{Research Data and Setup}

For the experimental evaluation, we use three publicly  available datasets. 
The \textbf{Hate Speech} dataset provided by a recent Kaggle competition\footnote{\url{https://www.kaggle.com/c/jigsaw-toxic-comment-classification-challenge}} consists of Wikipedia comments manually labeled for toxic behavior.  
We unify different types of toxicity in the datset to a binary classification task (toxic / non-toxic) and run our experiments on a subset of random 40,000 comments. 
The \textbf{IMDB} dataset \cite{maas2011learning} consists of 50,000 highly polarized English film reviews, which either are associated with a positive or negative user sentiment. 
Finally, the \textbf{20NewsGroups}  dataset \cite{lang1995newsweeder} comprises 18,846 English documents which are grouped in 20 different news  topics. 
As training, test, and evaluation sets, we randomly sample from the Hate Speech, IMDB, and 20NewsGroups datasets and apply train-test-evaluation splits of 20,000:10,000:10,000, 25,000:12,500:12,500 and 9,846:4,500:4,500 respectively. We perform all experiments five times with randomized train-test data splits and a constant held-out evaluation set.

Moreover, we use three common NN architectures from the literature, further referred to as CNN, KimCNN, and DistilBERT. 
\textbf{CNN} consists of one convolutional layer, a global max pooling layer and two fully connected dense layers similar to recent studies on app reviews and tweets classification \cite{stanik2019class}.  
We apply Dropout before each weight layer with a rate of $0.4$ and use L2-Regularisation with a kernel penalty of 1e-05.
As word representations, we take 100 dimensional trainable vectors which are randomly initialized.
For \textbf{KimCNN}, we follow the NN architecture and configuration suggested by \citet{kim2014convolutional}. The author proposes a multichannel convolutional NN with different filter region sizes, followed by a 1-max pooling layer.  As word representations, we take static 300 dimensional Google word2vec embeddings which are pretrained on 100 billion news articles \cite{mikolov2013distributed}. 
Lastly, we use the popular state-of-the-art text classification approach \textbf{DistilBERT} \cite{sanh2019distilbert}, a distilled version of the Bidirectional Encoder Representations from Transformers (BERT) \cite{devlin2018bert}. DistilBERT consists of 40\% fewer parameters compared to BERT and is thus much more efficient to train while retaining about 97\% of its performance.
We fine-tune DistilBERT via the default settings of the Huggingface Trainer API.\footnote{\url{https://huggingface.co/transformers/}} 

We apply the three uncertainty modeling techniques MCD, BBB, and Ensemble (described above) to each of the classifiers.
For MCD and BBB,  $50$ stochastic forward passes are applied. We use $5$ NNs as the size of our ensemble. It has been shown that larger ensembles do not significantly improve uncertainty estimations \cite{lakshminarayanan2017simple}.
For the MCD and Baseline approach, we perform inference on the same trained model since they share the same training procedure.
For DistilBERT we only implement MCD by activating the model's internal Dropout layer at inference time as performed by \citet{miok2021ban}. 
For the implementation of BBB into CNN and KimCNN we exchange the network's layers with Bayesian layers using the TensorFlow Probability library.\footnote{\url{https://www.tensorflow.org/probability}} BBB cannot be directly applied to DistilBERT. This would require altering the network's architecture and retraining the model from scratch.
%
Finally, we use the Kneedle algorithm \cite{satapaa2011knee} to detect the point of saturation as discussed in Section \ref{sec:saturation}. Since real saturation curves are usually not smooth we use polynomial interpolation to fit a spline used for detecting saturation points. 
Our replication package is publicly available online.\footnote{\url{https://github.com/jsandersen/CMT}}

\begin{figure*}[t]
	\centering
	\includegraphics[width=0.329\textwidth]{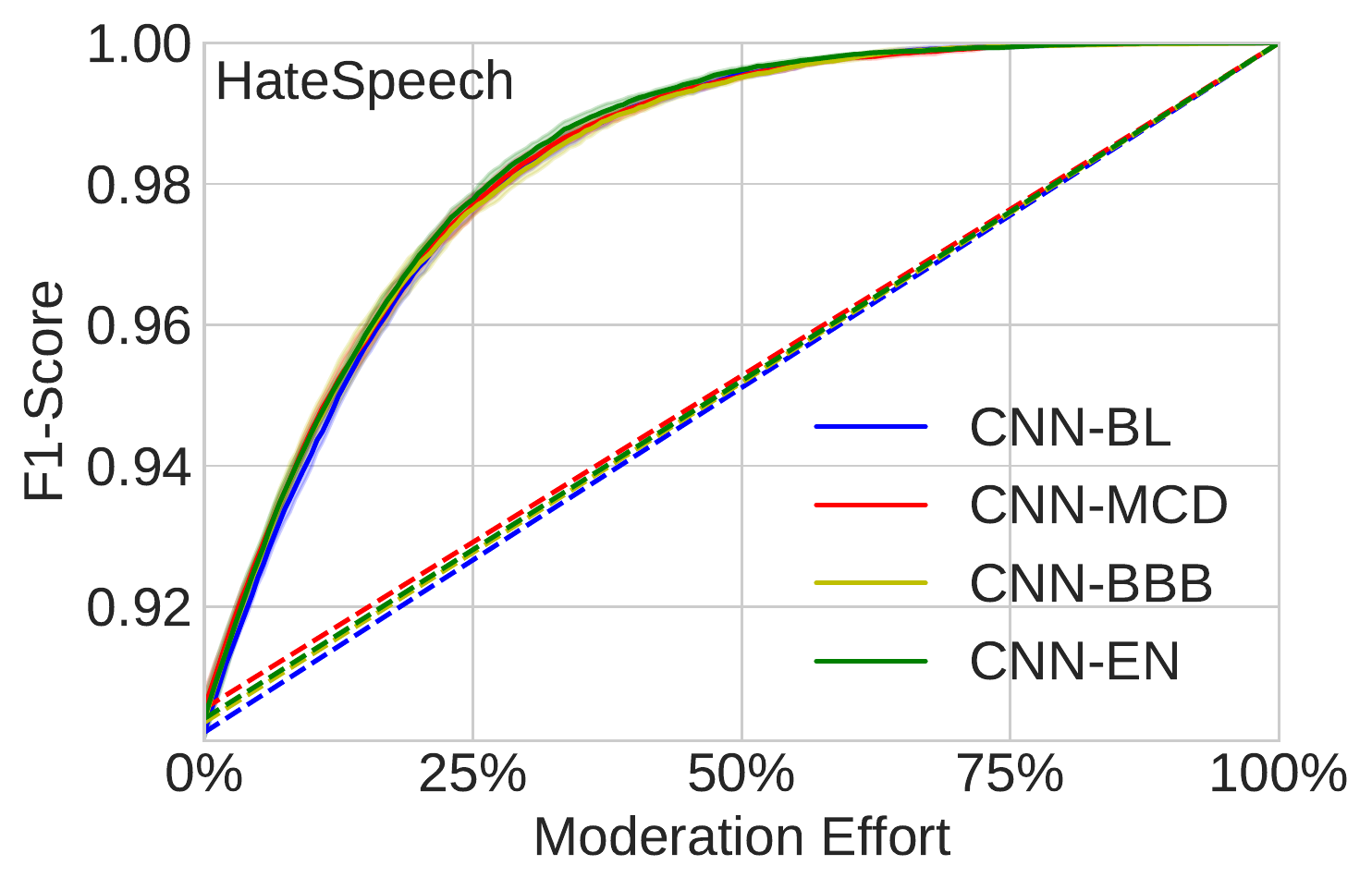}
	\includegraphics[width=0.329\textwidth]{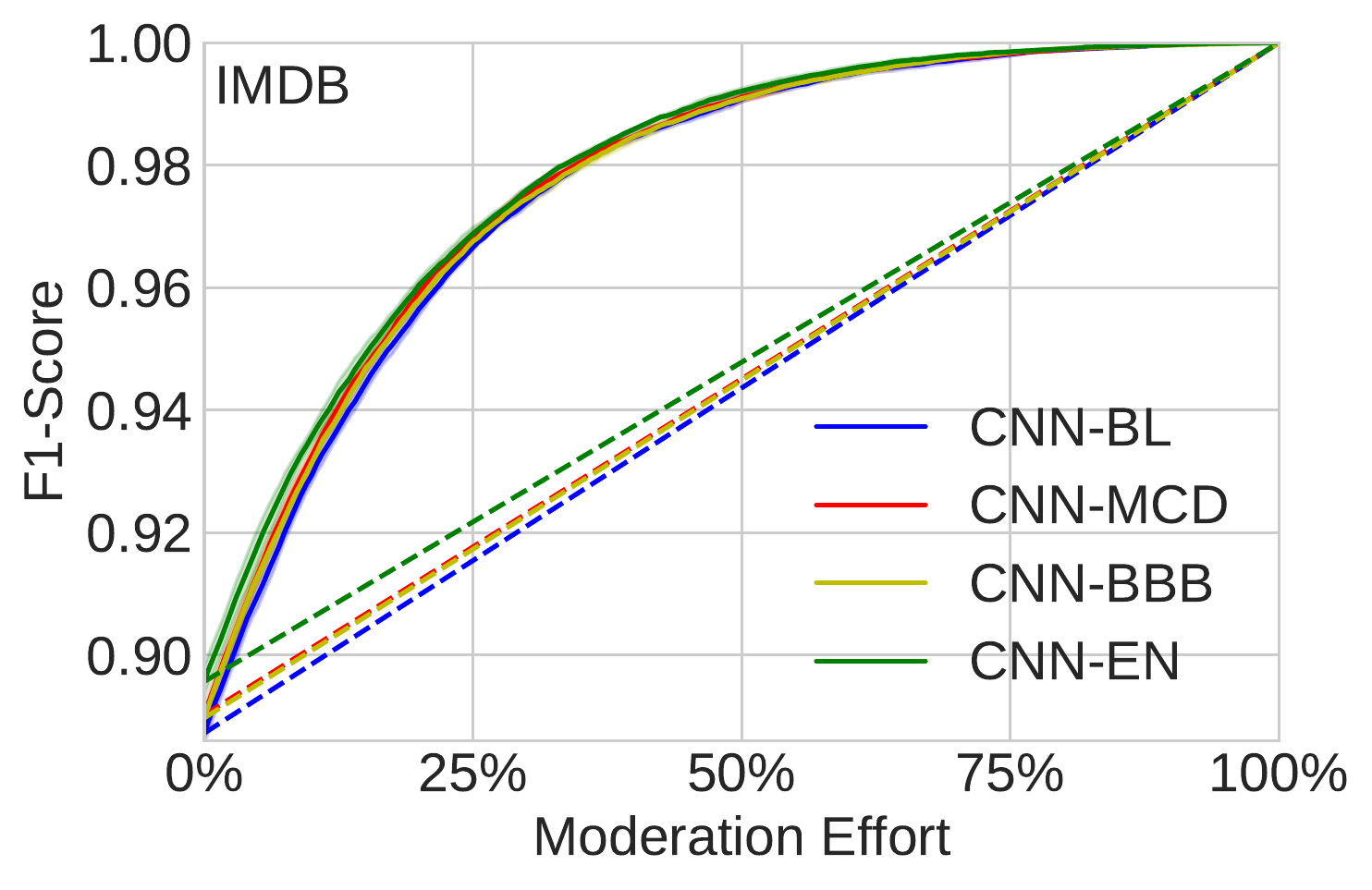}
	\includegraphics[width=0.329\textwidth]{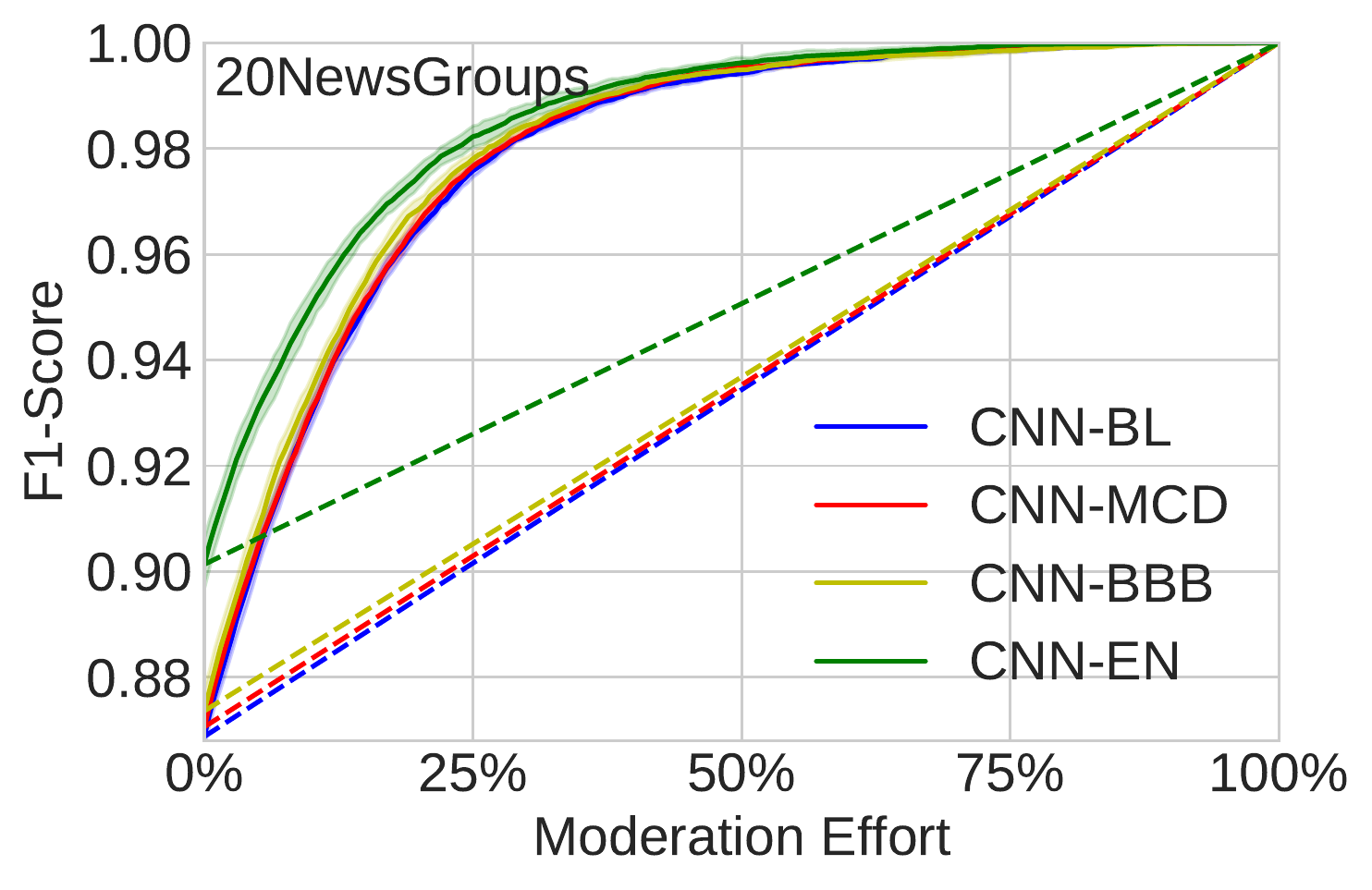}
	
	\includegraphics[width=0.329\textwidth]{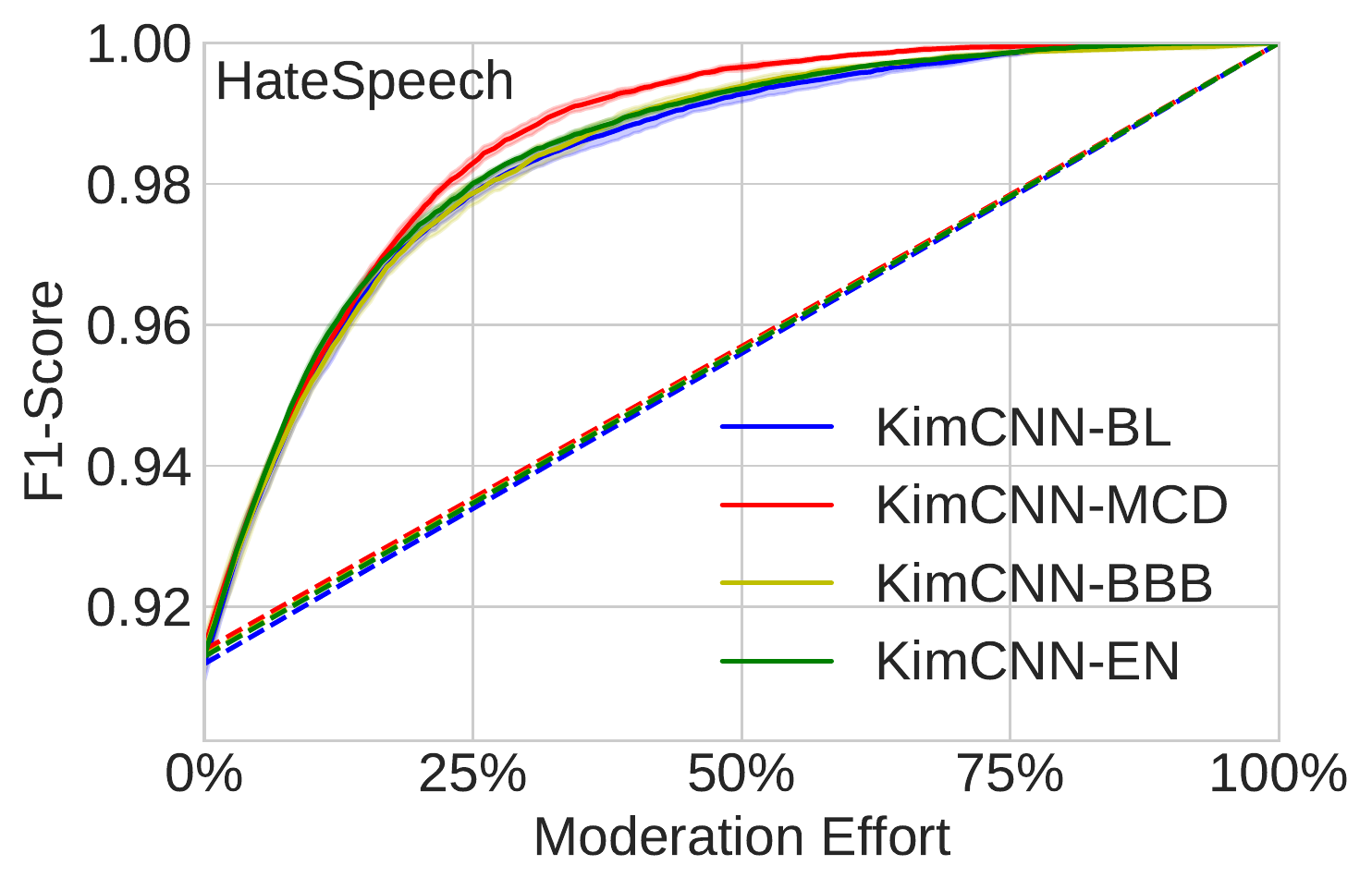}
	\includegraphics[width=0.329\textwidth]{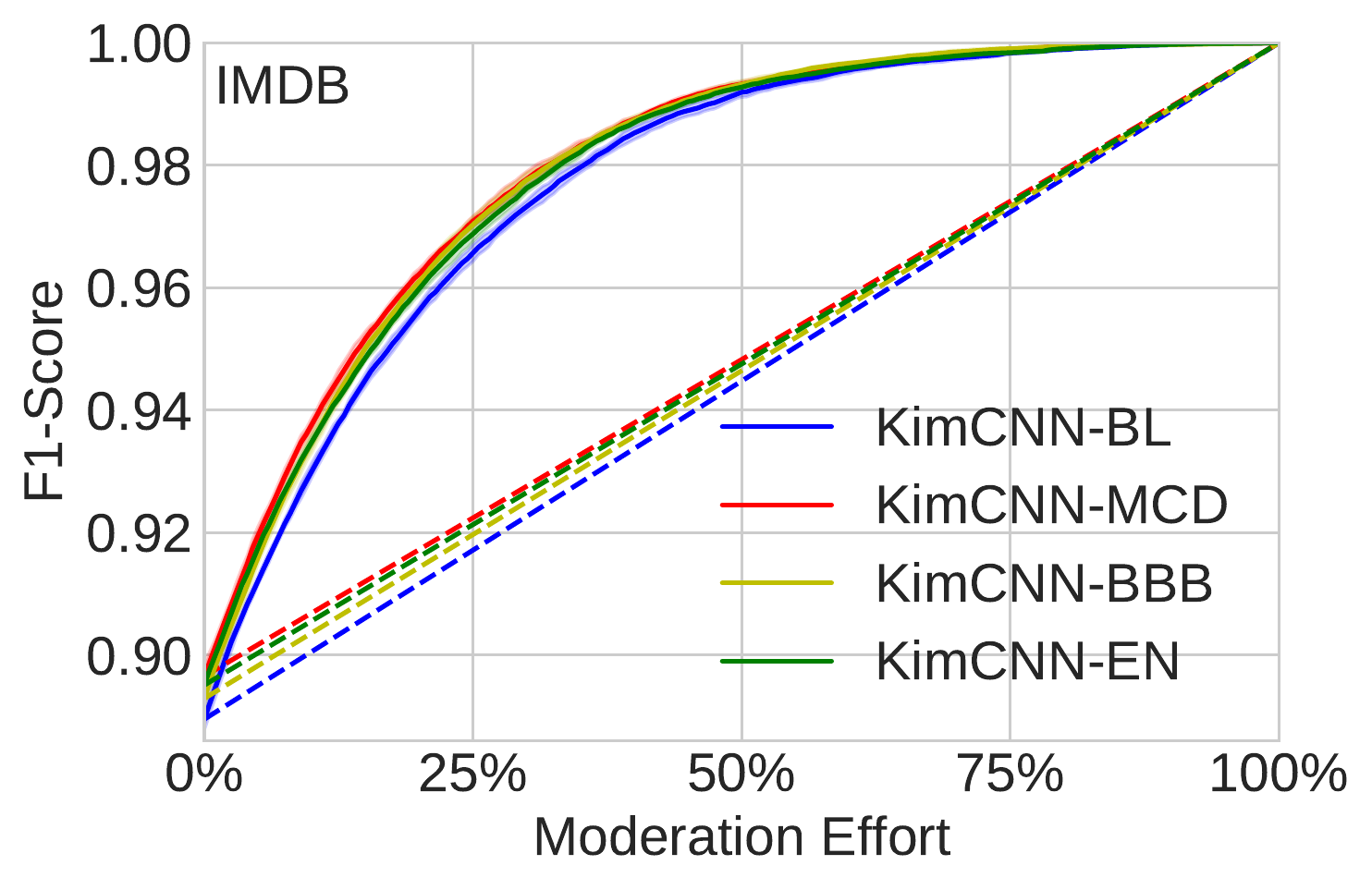}
	\includegraphics[width=0.329\textwidth]{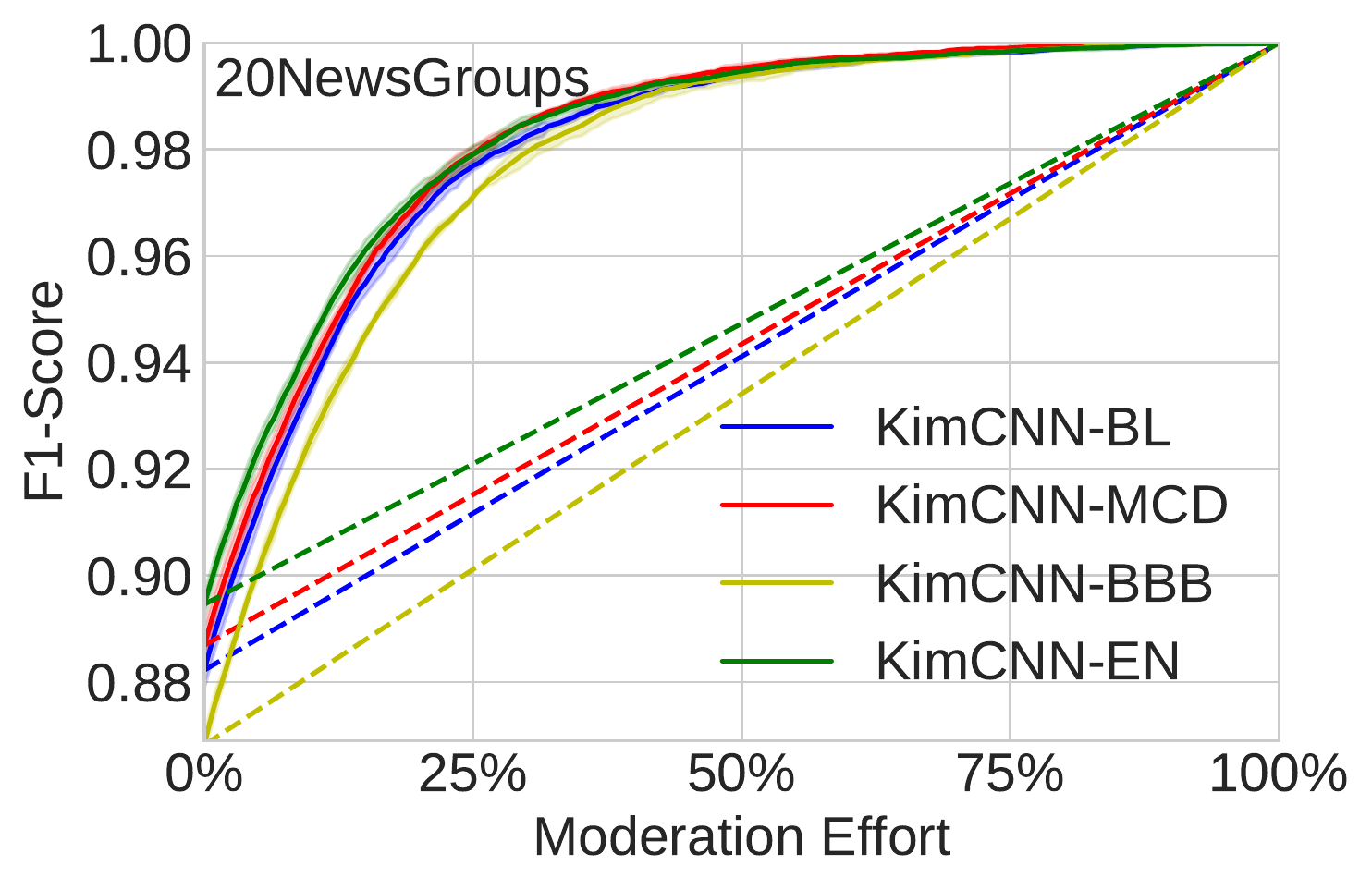}
	
	\includegraphics[width=0.329\textwidth]{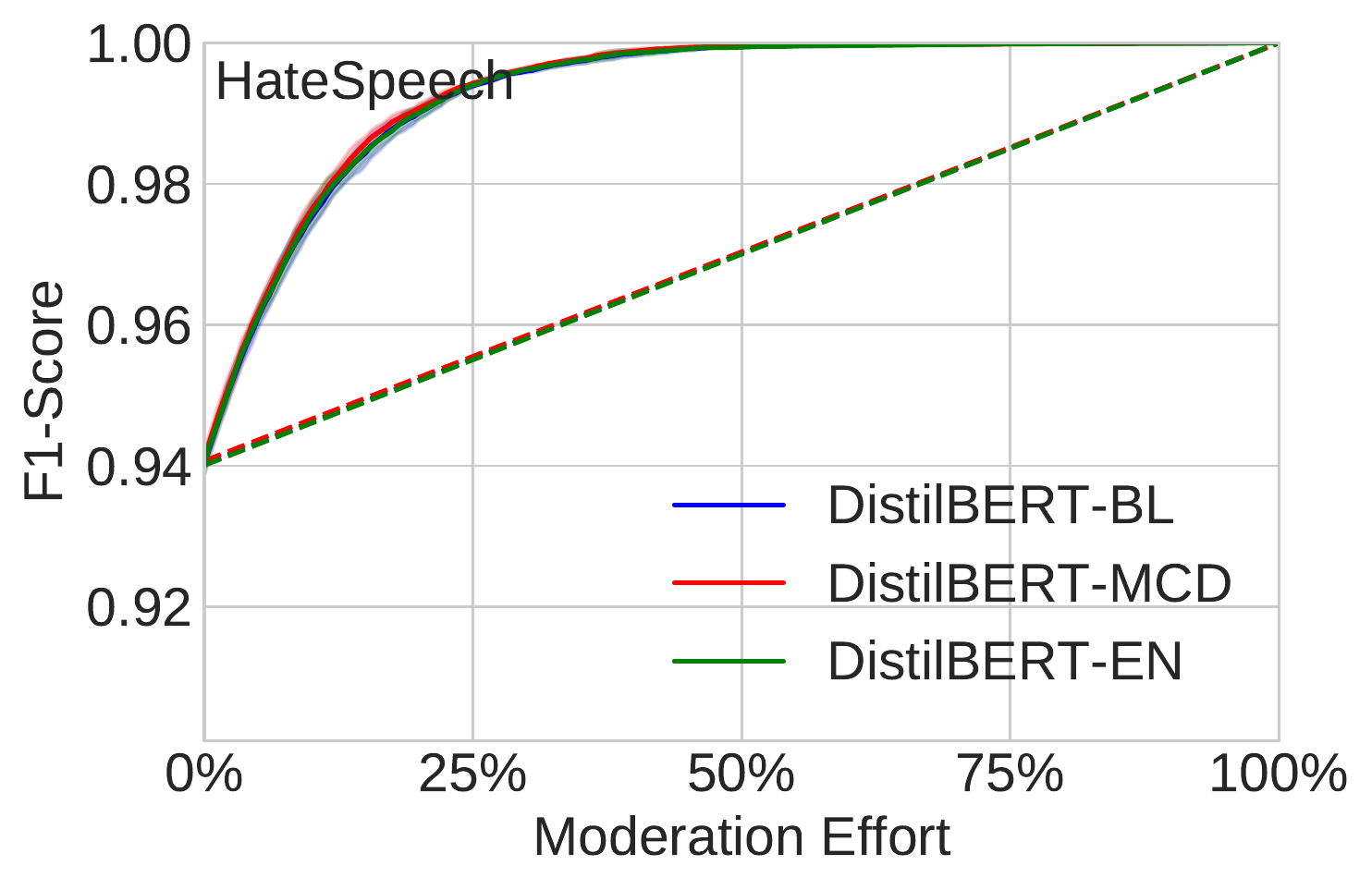}
	\includegraphics[width=0.329\textwidth]{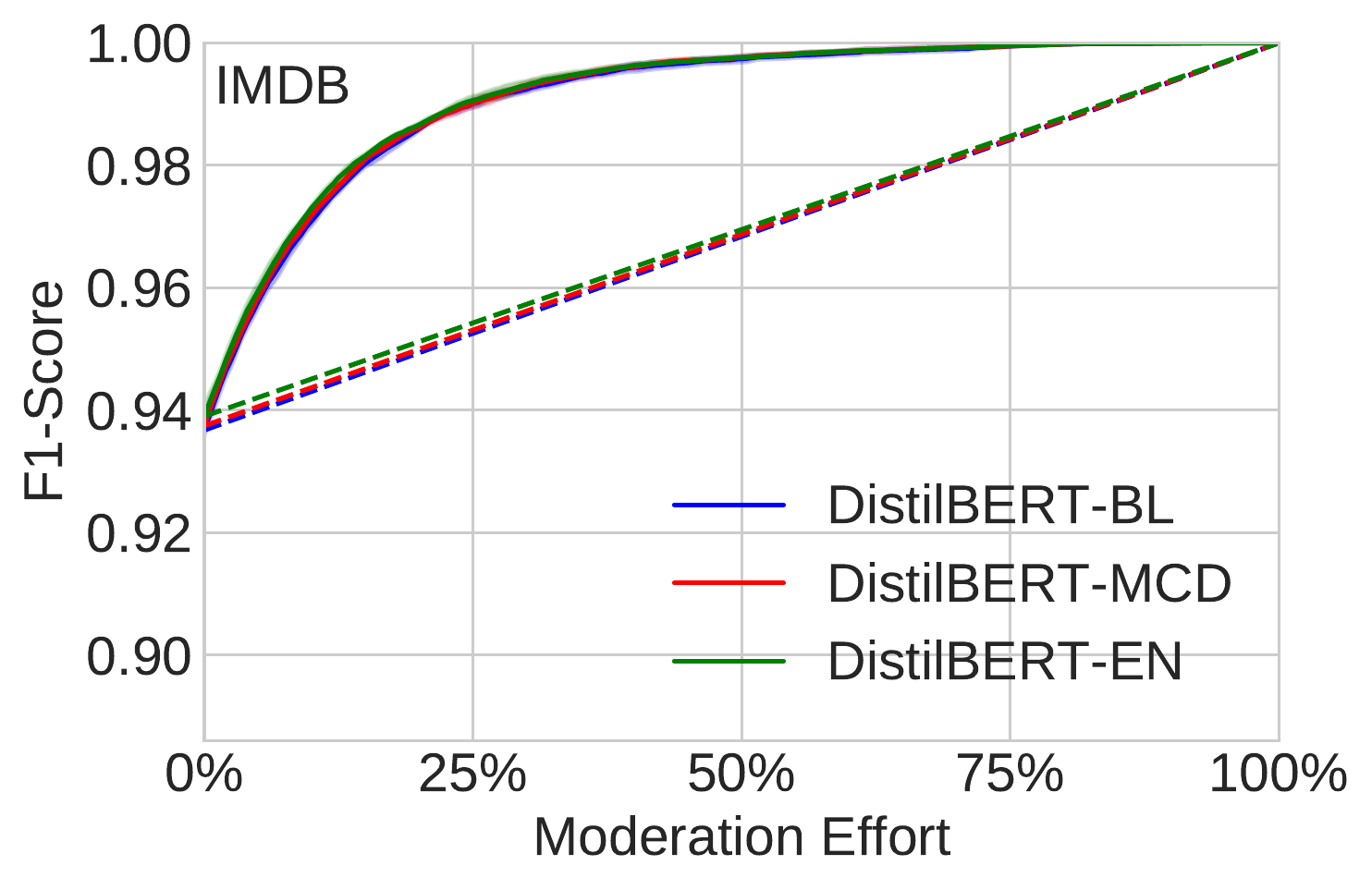}
	\includegraphics[width=0.329\textwidth]{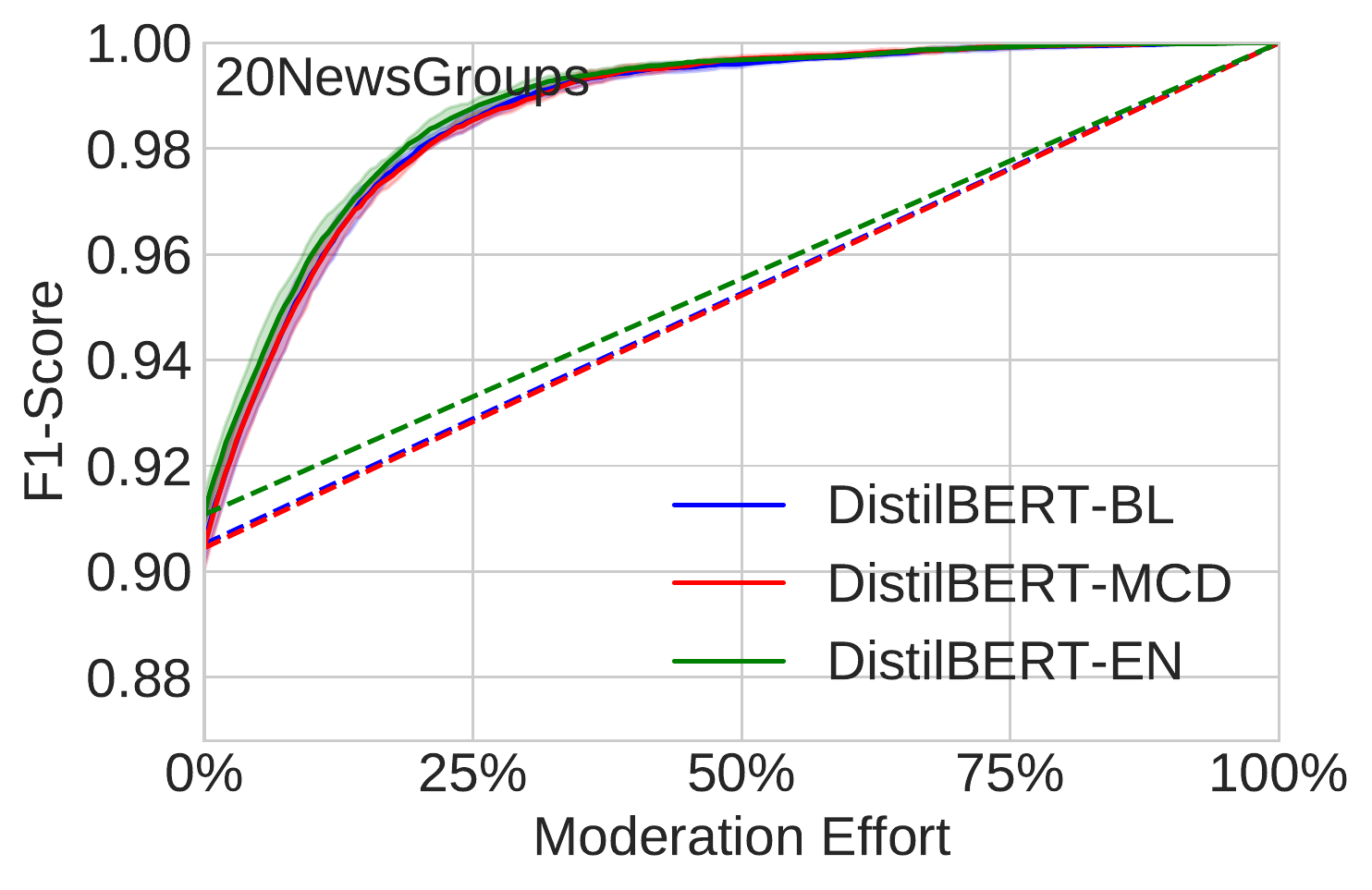}
	
	\caption{Accuracy gain for the three considered datasets \textbf{with the proposed moderation} using different uncertainty modeling and the Baseline (BL). Dotted-lines illustrate the F1-scores of a random moderation strategy.}
	\label{figure:efficency}	
\end{figure*}

\section{Experiments Results}
\label{Results}

\subsection{Extending Classifiers with Uncertainty Modeling}
\label{Distribution of Misclassification}

Table \ref{Performance Overview} presents the initial performance of the classifiers when no manual moderation is performed.
The table is organized in evaluation metrics and uncertainty modeling techniques which are applied to each dataset and classifier. 
Each cell consists of the mean followed by the standard deviation of five independent classification runs. 
For each classifier and dataset the scores of the best performing uncertainty modeling techniques are highlighted in green and the lowest scores in red.
\textit{Mean Conf.} represents the mean confidence score of all misclassified \textit{(Mis.)} and successful classifications \textit{(Suc.)}. The mean confidence range (\textit{Range}) is computed as the range between the mean confidences.

The results reveal  that CNN and KimCNN reach similar F1-score across all experiments ranging from 86.9 to 91.3\%.  
DistilBERT performs as expected better, particularly for binary classification (Hate Speech and IMDB) with an F1-score of up to 94.1\%. 
However, DistilBERT also leaves room for improvements still being away from a top F1-score around 99\%.  
The results show that across all experiments the explicit modeling of uncertainty only has a small effect compared to the F1-scores of the Baselines (less than 1\%). 
Only an Ensemble on CNN and 20NewsGroups  (multi label classification) attains an improvement of 3.2\% reaching $\sim$90\%.
MCD performs best on the Hate Speech dataset whereas an Ensemble performs best on the 20NewsGroups and IMDB datasets. BBB never reaches the overall highest F1-score.

The confidence scores reveal that an explicit modeling of uncertainty implies less overconfident wrong outputs compared to the Baseline. 
The uncertainty modeling does increase the confidence range between successful and misclassified classifications. 
KimCNN provided the least overconfident wrong outcomes followed by DistilBERT.
Our results also indicate that MCD, BBB, and an Ensemble generally outperform the Baseline in terms of misclassification detection (AUC-ROC). 
No specific technique consistently outperforms the others.

\subsection{Moderated Classification}
\label{Moderated Classifier Performance}

\begin{table*}[t]
	\centering
	\caption{Top F1-scores and moderation load (in \%) achieved using our saturation-based moderation strategy.
	}
	\resizebox{1\textwidth}{!}{%
		\begin{tabular}{c|c|c|c|c||c|c|c|c||c|c|c|c|c|}
			\cline{2-13}
			\multicolumn{1}{l|}{} & \multicolumn{4}{c||}{\textbf{Hate Speech}} & \multicolumn{4}{c||}{\textbf{IMDB}} & \multicolumn{4}{c|}{\textbf{20NewsGroups}}  \\ \cline{1-13}
			\multicolumn{1}{|c||}{\textbf{Saturation}}    & Baseline & MCD & BBB & Ensemble & Baseline & MCD & BBB & Ensemble & Baseline & MCD & BBB & Ensemble  \\ \cline{1-13}
			
			\multicolumn{8}{l}{} \\[-0.7ex] \hline
			
			\multicolumn{1}{|c||}{Moderation Load}   &  
			31.7&  31.6&  31.2&  31.8 &
			33.2 &  32.2 &   32.9 &  33.8 & 
			28.5 &   28.6 &  27.5 &  27.7 & 
			\parbox[t]{2mm}{\multirow{3}{*}{\rotatebox[origin=c]{90}{\textbf{\small CNN}}}}\\ \cline{1-13}
			\multicolumn{1}{|c||}{F1-score}   
			& \begin{tabular}[c]{@{}l@{}}98.08\\(+7.9)\end{tabular}  
			&  \begin{tabular}[c]{@{}l@{}}98.10\\(+7.5)\end{tabular} 
			&  \begin{tabular}[c]{@{}l@{}}97.94\\(+7.6)\end{tabular} 
			&  \begin{tabular}[c]{@{}l@{}}98.25\\(+7.8)\end{tabular} 
			
			&\begin{tabular}[c]{@{}l@{}}97.80\\(+9.1) \end{tabular} 
			&\begin{tabular}[c]{@{}l@{}}97.77\\ (+8.8)\end{tabular}   
			& \begin{tabular}[c]{@{}l@{}}97.77\\(+8.8)\end{tabular}   &\begin{tabular}[c]{@{}l@{}}98.03\\(+8.4)\end{tabular} & \begin{tabular}[c]{@{}l@{}}98.10\\(+11.2)\end{tabular} & \begin{tabular}[c]{@{}l@{}}98.15\\(+11.1)\end{tabular} &\begin{tabular}[c]{@{}l@{}}98.13\\(+10.8)\end{tabular}   
			&\begin{tabular}[c]{@{}l@{}}98.46\\(+8.3)\end{tabular}  
			
			&\\ \cline{1-14}	
			
			\multicolumn{8}{l}{} \\[-0.7ex] \hline
			
			\multicolumn{1}{|c||}{Moderation Load}   &  
			26.6& 29.1& 28.7& 26.2&
			35.5  &  33.4 &  33.1 &  34.9&  
			27.8&  28.2 &   30.7 &  27.7 & 
			\parbox[t]{2mm}{\multirow{3}{*}{\rotatebox[origin=c]{90}{\textbf{\small K.CNN}}}}\\ \cline{1-13}
			\multicolumn{1}{|c||}{F1-score}   
			&  \begin{tabular}[c]{@{}l@{}}97.57\\(+6.4)\end{tabular} 
			&\begin{tabular}[c]{@{}l@{}}98.39\\(+7.0)\end{tabular} 
			&\begin{tabular}[c]{@{}l@{}}97.77\\(+6.5)\end{tabular} 
			&\begin{tabular}[c]{@{}l@{}}97.70\\(+6.4)\end{tabular} 
			&  \begin{tabular}[c]{@{}l@{}}98.01\\(+9.1)\end{tabular} 
			&\begin{tabular}[c]{@{}l@{}}98.13\\(+8.5)\end{tabular}
			&\begin{tabular}[c]{@{}l@{}}98.12\\(+8.8)\end{tabular}   
			&\begin{tabular}[c]{@{}l@{}}98.20\\(+8.7)\end{tabular}  
			& \begin{tabular}[c]{@{}l@{}}98.00\\(+9.8)\end{tabular} 
			&\begin{tabular}[c]{@{}l@{}}98.32\\(+9.6)\end{tabular}
			& \begin{tabular}[c]{@{}l@{}}98.04\\(+11.2)\end{tabular} 
			&\begin{tabular}[c]{@{}l@{}}98.24\\(+8.8)\end{tabular}  
			
			& \\ \cline{1-14}			
			
			\multicolumn{8}{l}{} \\[-0.7ex] \hline
			
			\multicolumn{1}{|c||}{Moderation Load}   & 
			24.5 & 23.9 & -  &   24.5 & 
			25.3&   25.1 & -  &  24.6 &  
			25.5 &   25.9 & -  & 25.5  & 
			
			\parbox[t]{2mm}{\multirow{3}{*}{\rotatebox[origin=c]{90}{\textbf{\small D.BERT}}}}\\ \cline{1-13}
			\multicolumn{1}{|c||}{F1-score}  
			&   \begin{tabular}[c]{@{}l@{}}99.36\\ (+5.3)\end{tabular}
			& \begin{tabular}[c]{@{}l@{}}99.37\\ (+5.3)\end{tabular}
			& - 
			&   \begin{tabular}[c]{@{}l@{}}99.37\\ (+5.4)\end{tabular} 
			
			&   \begin{tabular}[c]{@{}l@{}}99.03\\ (+5.4)\end{tabular}
			&   \begin{tabular}[c]{@{}l@{}}99.01\\ (+5.3)\end{tabular}&  
			- 
			&   \begin{tabular}[c]{@{}l@{}}99.04 \\ (+5.1)\end{tabular}
			&    \begin{tabular}[c]{@{}l@{}}98.60\\ (+8.1)\end{tabular}
			&   \begin{tabular}[c]{@{}l@{}}98.62\\ (+8.2)\end{tabular}
			&  - 
			&   \begin{tabular}[c]{@{}l@{}}98.82\\ (+7.7)\end{tabular} 
			
			&\\ \cline{1-14}		
			
		\end{tabular}
	}
	
	\label{table:saturation}
\end{table*}

Next, we investigate the overall F1-scores of our moderated classifier framework. Figure \ref{figure:efficency} shows  the F1-scores for the Hate Speech, IMDB, and 20NewsGroups datasets. The y-axis plots the F1-score and the x-axis indicates the corresponding  manual moderation effort. The F1-score considers  artificial classification outcomes as well as the manual labeled  examples. In our experiments, manual labeling  is done by picking the ground truth label.
The \textit{Least Confident} score function\footnote{$unc_{LC}[y|x,D] := 1 - \max_c p(y=c|x, D)$} is used for all experiments as it reaches the highest F1-score in most cases and performs most consistently.

The accuracy gains with our framework depicted in Figure \ref{figure:efficency} reveal a significant accuracy increase  compared to a random moderation strategy, which is depicted by the dotted lines. Furthermore, as expected, the moderation becomes less efficient with an increasing moderation load, as misclassifications are more common when a classifier reports large uncertainty scores.

All studied classifiers show a similar moderation behavior. 
All accuracy curves follow the shape of a saturation curve assumed in Section \ref{Moderated Classifier}.
The highest variations occur on the 20NewsGroups dataset. Furthermore, the difference between all approaches becomes less with an increasing moderation effort. 
By moderating more instances manually, more similar F1-scores are reached by all classifiers.
Overall, an Ensemble of homogeneous NNs and MCD reaches the overall highest F1-score with the least moderation effort (and BBB slightly less). On average, the Baseline requires slightly more manual effort to reach the same F1-score.

As the accuracy gains decrease with an increasing moderation effort, we calculate saturation points to stop the moderation before it becomes inefficient.  Table \ref{table:saturation} lists those saturation points for the LC score function. 
The absolute improvement of the F1-score is shown in brackets. The table shows that our moderation approach is able to achieve a F1-score of 98 to 99\% on all classification tasks while maintaining an efficient human moderation. 
These F1-scores can be achieved with all evaluated classifiers and uncertainty estimation techniques.
Saturation points are reached after moderating  $\leq$33.1\% of the dataset using CNN and KimCNN and $\leq$25.5\% using DistilBERT.
All classifiers provide a similar trade-off between achieved F1-score and moderation effort. 
However, the Baseline is not optimal since it either saturates with slightly higher moderation efforts or provides a lower F1-score compared to MCD, BBB, and an Ensemble.
On IMDB the MCD reaches saturation with the least moderation effort, while providing a high level of accuracy. 
On 20NewsGroups an Ensemble performs slightly better. 

Interestingly, DistilBERT requires the least manual effort while achieving the highest level of accuracy, i.e., a F1-score of $ 98.6 - 99.37\%$.
The results also reveal that models with a low initial F1-score reach higher absolute F1-score improvements.  
Overall, using our framework, a moderator has to label up to 73.3\% (Hate Speech), 71.0\% (IMDB) and 70.9\% (20NewsGroups) \textit{less data} instances compared to a random moderation strategy. 
Based on these results, we conclude the answers to our research questions: 

\paragraph{Answer RQ1} 
Explicitly modeling uncertainties of NN classifiers only has a minor impact on the accuracy compared to the baseline.
All techniques provide similar F1-score improvements when a human is moderating a certain number of classification outcomes. 
Only on a multi-class classification problem (20NewsGroups) an Ensemble provides slightly better ($\geq$1\%) F1-scores compared to a traditional deterministic NN (Baseline). 
Overall, a five NN Ensemble and MCD achieves the best accuracy improvements compared to the Baseline and BBB.
The moderated classifiers performed  best with the Least Confident score function.

\paragraph{Answer RQ2}
Moderating the outcomes of classifiers can lead to top F1-scores between 97 and 99\% using both rather weak (CNN / KimCNN) and strong (DistilBERT) classifiers while efficiently limiting human involvement. 
Using DistilBERT an absolute F1-score improvement of +5.3 (Hate Speech) +5.1 (IMDB) and +7.7\% (20NewsGroups) is reached by moderating $\leq$25.5\% of the data. 
A saturation-based moderation saves up-to 73.3\% (Hate Speech) 71.0\% (IMDB) and 70.9\% (20NewsGroups) of effort compared to a random moderation to reach the same F1-score.

\section{Discussion}

\label{dicussion}
\subsection{Implications}

Our results indicate that the mere explicit uncertainty modeling can barely enhance the accuracy of automatic text classifiers.
However, our  moderation framework can substantially improve the accuracy and thus the acceptance of rather weak (CNN / KimCNN) as well as strong (DistilBERT) classifiers. Compared to cases where classification decisions have to be done fully manually when the accuracy of automatic classifiers is inapplicable in practice, our semi-automated framework would require one fourth to one third of the manual effort to obtain a top accuracy of $\sim$98-99\%. This is a substantial saving of resources. 
The major advantage of the framework is the ending of the moderation when it becomes inefficient. 
Further, our results indicate that the framework also works  well with usual NNs (Baseline). Even if the accuracy improvement and effort minimization are not as good as with explicit uncertainty modeling, the Baseline reaches similar top F1-scores.
Thus, the cost of implementing and adopting the framework to existing  classifiers is rather limited. Additional costs of explicit uncertainty modeling should  be assessed against the marginal achievable improvements.

Clearly, the usefulness of our framework depends on the application scenario at hand. 
In particular, it is crucial to first investigate:
\begin{itemize}
    \item Whether a top accuracy, of e.g.~99\% is expected by users or not.
    \item Whether and how human moderation is applicable, and if the moderation can be trusted. 
    \item Whether the goal of maximizing the accuracy while minimizing the human effort is desired. 
\end{itemize}
We think that semi-automated approaches are particularly important in domains with a very large number of classifications and where classification mistakes are  costly, for instance when user comments need to be moderated in a public debate space such as comments in news outlets \cite{loosen2018making,boberg2018moral} or in Wikipedia as in the Hate Speech dataset. 
A pure automated classification and analysis of inherently ambiguous text, e.g.~reflecting human opinions or outlining novel ideas will quickly reach its limits. Even humans might not totally agree on a uniform labeling of complex texts \cite{guillermocarbonell2016measuring}. 
As shown in our experiments, most documents can accurately be labeled  by a machine and do not require human effort. 
However, complex or ambiguous texts might not be handled appropriately by black-and-white categorization and machines might be unable to make reliable classifications. 
By placing a human in the loop, human creativity and reasoning contribute to efficiently solving such difficult tasks.

With moderation, additional data is continuously collected and can be used to re-train the classifier from time to time. Re-training often prevents an accuracy decay of the underlying classifier over time \cite{moreno2012unifying} and can (but do not necessarily) improve its accuracy  \cite{arnt2003learning}. 
Our results suggest that the moderation  would benefit from a higher initial accuracy as the amount of human involvement seems lowest here. 
Further research is required to better understand the interplay between in-operation moderation and active learning.

\subsection{Limitations}

Our moderation framework builds on the  assumption that a classifier's expected accuracy curve can be used to estimate how it would perform in operation.
This assumption depends on whether the dataset used for the evaluation and the derived saturation point represent the real data distribution. In cases where this assumption is not reasonable and the real data distribution is significantly different (e.g. due to particular events), more research would be needed e.g. to monitor and potentially adjust the operational data distribution and the saturation curve accordingly.

In our experiments we assume that human moderators do not commit errors. While a flawless moderator is generally assumed in the review of interactive machine learning approaches like active learning \cite{burkhardt2018semi,gal2017deep,houlsby2011bayesian}, the assumption does not have to apply to all real scenarios \cite{sheng2008get}. 
Generally, as already discussed above, annotations from domain experts are seen as more trustful than machines, especially on difficult tasks such classification of ambiguous text. 
Human annotations are often considered as the ground truth for classification tasks and are used to initially train a classifier \cite{lewis1994sequential}. Therefore, we assume that domain experts annotate instances more reliably than machines in real-world domains. This assumption may have limitations in practice, as people may make mistakes too.

Interactive ML approaches such as our framework are confronted with the limitation of scalability. Even a small fraction of human involvement can lead to an enormous manual effort when the data to be classified is very large. Our framework can limit human involvement to  $23.9-25.4\%$ of the data assessment in order to reach a top F1-score. Finally, human moderators have to decide whether spending these efforts is applicable and desired.

Our approach also faces the limitation that uncertainty estimation approaches are unable to identify highly certain classifications which are actually wrong (unknown-unknowns) \cite{attenberg2011beat}. 
Thus, it is unrealistic to avoid all misclassifications without manually checking all the data. 
However, we have shown that the majority  of misclassifications can be efficiently identified  by our approach leading to high F1-scores $\sim$98-99\%. 

Finally, as for every empirical evaluation, our results are dependent on the datasets, metrics and setting used. While we refrain from claiming the generalizability of the concrete quantitative results, the diversity of the datasets and classification models used give us enough confidence on the general observed trends for text classification. For other classification tasks, a replication using other datasets and model architectures would be required. 

\section{Related Work}
\label{Related Work}

The moderation of classifier outputs can clearly be considered as an application of the Human-in-the-Loop (HiL) paradigm \cite{holzinger2016interactive}. So far, most HiL implementations focus on querying additional labels from humans for the purpose of training in order to reduce a classifier's uncertainty, commonly referred to as active learning \cite{settles1995active}. In comparison, our framework aims to efficiently prevent error-prone classifications during operation and thus to further enhance the accuracy of an already trained classifier.

Different approaches have been previously discussed to coordinate human involvement in semi-automatic text classification. 
To our best knowledge, we are the first to investigate the \textit{human efficient}, moderation of text classifiers.
\citet{pavlopoulos2017deep} propose to search for confidence thresholds which maximize a classifier's accuracy, when classification outcomes are moderated manually. However, this approach requires moderators to  set the amount of data they are willing to moderate. Assessment of the efficiency is not made.  
Another approach to reduce error prone classifications is to let classifiers abstain when no clear decisions can be made \cite{cortes2016learning,ramaswamy2018consistent}. This can be performed by adding an additional label to the classification task or by training a separate and independent classifier. Abstained instances could be, similarly to our approach, passed to human moderators. Our work focuses on  uncertainty modeling using NNs. 
\citet{geifman2017selective} propose a classification approach with a reject option which additionally allows practitioners to set a desired  level of risk. Similar to our approach, they aim to ensure a certain classification performance. In contrast, they do not focus on the efficiency of human involvement.

\citet{lee2018training} propose an approach to detect whether an instance is out-of the distribution of the training dataset and thus probably wrongly classified. However, their approach requires an auxiliary dataset representing out-of-distribution samples during training which is difficult to create.
\citet{de2021classification} introduce a semi-automated approach which directly optimizes a classifier for different automation levels. However, their approach is only applicable to convex-margin based classifiers and not to NNs.
\citet{xiao2021self} suggest a self-checking mechanism for NN, where the features of the internal layers  are  used  to  check the reliability of predictions. In contrast, our approach uses predictive uncertainties obtained via a softmax function, which is rather simple to implement. 

Moderating classifiers' outcomes is also related to the field of explainable ML, in particular explaining individual classification outcomes \cite{ribeiro2016should}. Studies indicate that explaining relevant words of a class outcome support human annotation tasks by, e.g., reducing the annotation time needed per instance and increasing user trust \cite{vsvec2018improving,ribeiro2016should}. Our approach is likely to benefit from explaining artificial decision-making as well as model uncertainties \cite{andersen2020word} during the moderation process.

\section{Conclusion}
\label{end}

This paper contributes to the Human-in-the-Loop AI paradigm. We particularly present a rather simple, semi-automated text classification framework to efficiently minimize unreliable and error-prone classification outcomes. Based on explicit uncertainty modeling, the framework seeks to prevent unconfident classifications by consulting human moderators. 
At its core, the framework uses a saturation-based moderation strategy, which limits the moderation load and keeps it human-resource-efficient. We conduct several benchmarking experiments including state-of-the-art classifiers and public datasets to examine the effectiveness of the suggested moderated classification.

Our evaluation shows that a moderated classifier can achieve a major increase in accuracy while limiting the moderation efficiency. 
With moderation, the F1-score of a convolutional NN for hate speech detection increases from initially 90.6 to 98.10\% limiting the manual effort to 31.6\%. Using DistilBERT, the benefits of a moderated classifier seems even stronger. 
Here, our framework accomplishes an improvement of the absolute F1-score from 94.1\% to 99.37\% while only moderating 23.9\% of the data. 
Across all our experiments, we increased the F1-score from initially $\sim$89-94\% to $\sim$98-99\% by manually moderating between a third to a fourth of the data. 
Our results indicate that an uncertainty-based moderated classification can increase the applicability and reliability of text classifiers, particularly in domains where a top accuracy of $\sim$99\% is required and a full manual classification  would be more expensive.

\section*{Acknowledgments}
The paper was supported by BWFGB Hamburg within the ``Forum 4.0'' project as part of the ahoi.digital funding line.

\balance

\bibliographystyle{acl_natbib}

\end{document}